\documentclass[10pt,twocolumn,letterpaper]{article}

\usepackage{iccv}              

\definecolor{iccvblue}{rgb}{0.21,0.49,0.74}
\usepackage[pagebackref,breaklinks,colorlinks,allcolors=iccvblue]{hyperref}
\usepackage{multirow}
\usepackage{notoccite}
\def\paperID{14648} 
\def\confName{ICCV}
\def\confYear{2025}

\title{DiT-VTON: Diffusion Transformer Framework for Unified Multi-Category Virtual Try-On and Virtual Try-All with Integrated Image Editing}

\author{
Qi Li\textsuperscript{1}\thanks{Equal contribution.} 
\quad
Shuwen Qiu\textsuperscript{2 *}\thanks{Work done during internship at Amazon.}
\quad
Julien Han\textsuperscript{1}\quad
Xingzi Xu\textsuperscript{1,3 \textdagger}\quad
Mehmet Saygin Seyfioglu\textsuperscript{1}\quad
Kee Kiat Koo\textsuperscript{1} \vspace{4pt}\\
Karim Bouyarmane\textsuperscript{1} \vspace{8pt}\\
\textsuperscript{1}Amazon \quad
\textsuperscript{2}University of California, Los Angeles (UCLA) \quad
\textsuperscript{3}Duke University
\vspace{8pt}\\
{\tt\small \{qlimz,hameng,xingzixu,mseyfiog,kiatkoo,bouykari\}@amazon.com}\\
{\tt\small xingzi.xu@duke.edu} \\
{\tt\small jantqiu@cs.ucla.edu} \\
{\color{red}{\tt\small \url{https://dit-vton.github.io/DiT-VTON/}}} \\
{\small \textbf{Work submitted in November 2024 to CVPR 2025}}
}

\begin{document}
  \maketitle
  \begin{abstract}
The rapid growth of e-commerce has intensified the demand for Virtual Try-On (VTO) technologies, enabling customers to realistically visualize products overlaid on their own images. Despite recent advances, existing VTO models face challenges with fine-grained detail preservation, robustness to real-world imagery, efficient sampling, image editing capabilities, and generalization across diverse product categories. In this paper, we present DiT-VTON, a novel VTO framework that leverages an architecture based on a Diffusion Transformer (DiT), renowned for its performance on text-conditioned image generation (text-to-image), adapted here for the image-conditioned VTO task. We systematically explore multiple DiT configurations, including in-context token concatenation, channel concatenation, and ControlNet integration, to determine the best setup for VTO image conditioning. 
To enhance robustness, we train the model on an expanded dataset encompassing varied backgrounds, unstructured references, and non-garment categories, demonstrating the benefits of data scaling for VTO adaptability. DiT-VTON also redefines the VTO task beyond garment try-on, offering a versatile Virtual Try-All (VTA) solution capable of handling a wide range of product categories and supporting advanced image editing functionalities, such as pose preservation, precise localized region editing and refinement, texture transfer and object-level customization. Experimental results show that our model surpasses state-of-the-art methods on public benchmark tests like VITON-HD, achieving superior detail preservation and robustness without reliance on additional image condition encoders. It also surpasses state-of-the-art models that have VTA and image editing capabilities on a varied dataset composed of thousands of product categories. As a result, DiT-VTON significantly advances VTO applicability in diverse real-world scenarios, enhancing both the realism and personalization of online shopping experiences. 
\end{abstract}    
  \section{Introduction}
\label{sec:intro}
\begin{figure*}[t]
 \centering
 \includegraphics[width=0.8\textwidth]{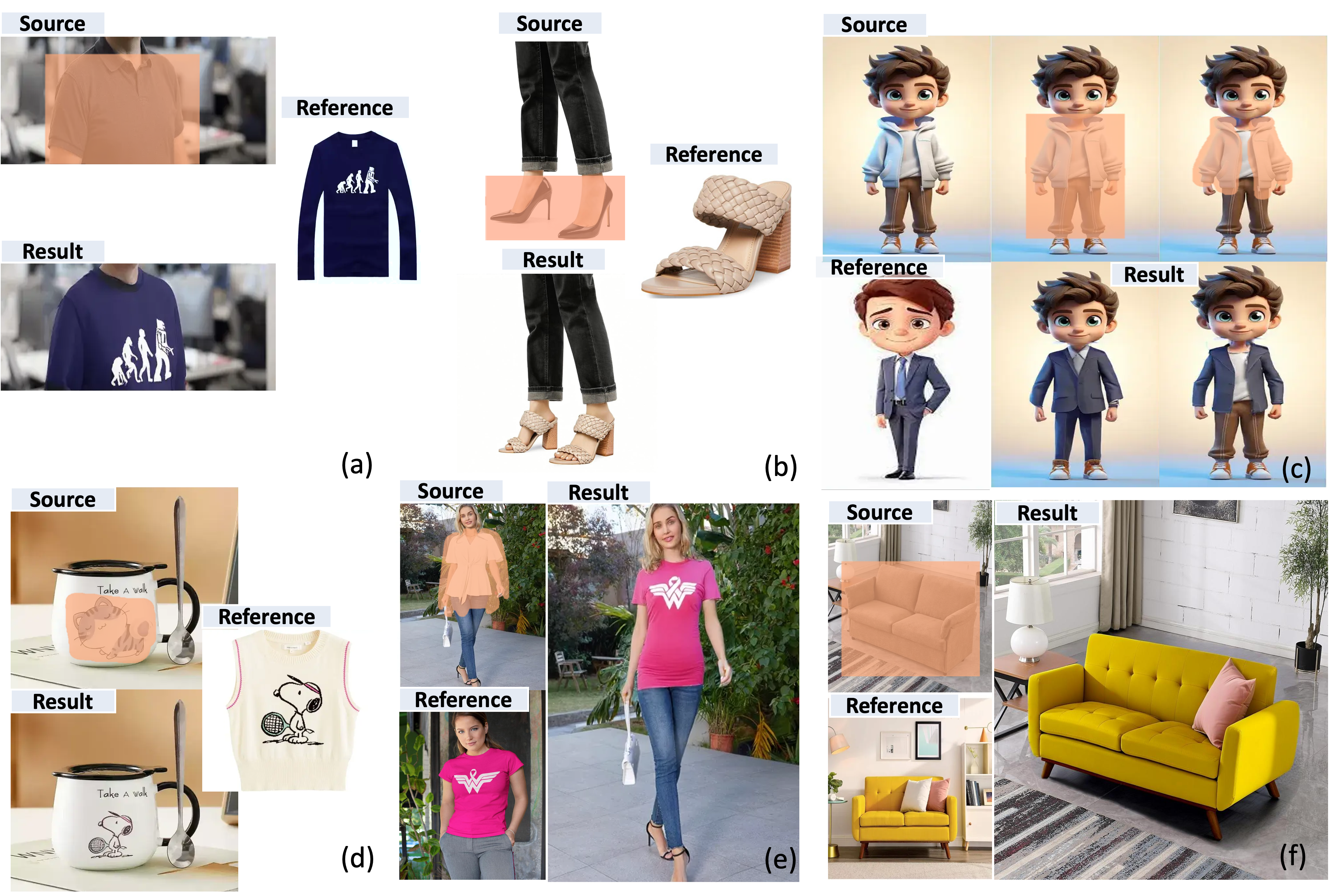}
 \caption{\textbf{Our DiT-VTON model applied to diverse use cases, enabling inpainting within user-specified editing regions using content guided by a reference image.} The model can semantically infer and generate expected objects, texture details, and even identify specific body parts for virtual try-on (VTO) tasks, showcasing its versatility in content-aware editing and synthesis. The model can also generate better product details and layouts for virtual try-all (VTA) tasks such as furniture. Please zoom in to see details preservation.}
 \label{fig:intro}
\end{figure*}
The rapid expansion of e-commerce has fueled demand for advanced tools that empower customers to virtually “try on” products by overlaying them onto their own images, enhancing both the personalization and interactivity of online shopping experiences. This technology, known as Virtual Try-On (VTO), requires a user’s image (source image), a specified mask indicating the intended placement of the product, and a product image (reference image). These components work together to seamlessly integrate the product into the designated area of the source image, creating a cohesive and realistic representation of the product on the user, as depicted in~\cref{fig:intro}. From a technical perspective, the VTO task can be formulated as an image-conditioned inpainting problem, wherein the challenge lies in filling masked regions in the source image with highly realistic, contextually accurate product visualizations.

Recent breakthroughs in U-Net-based Latent Diffusion Models (LDMs)~\cite{rombach2022high, esser2024scaling, nichol2021glide, saharia2022photorealistic} have propelled the capabilities of VTO models by enabling them to leverage pre-trained weights from text-to-image and inpainting diffusion frameworks. These advancements have led to state-of-the-art (SOTA) performance, marked by superior rendering quality and realism, broadening the potential for widespread application in the online retail space.

Despite substantial progress, existing SOTA VTO models  continue to encounter several key challenges that hinder their practical efficacy and versatility: \textbf{1. Fine-Grained Detail Preservation}. Maintaining fine-grained details (shown in~\cref{fig:intro}a), such as logos, text decorations, and intricate textures, remains a challenge. These small yet crucial elements occupy only a minor portion of the generated image area, often leading models trained with denoising score matching objectives to overlook them. Preserving these details is essential for delivering visually accurate representations of products. \textbf{2. Robustness to In-the-Wild Images}. Current models are predominantly trained on preprocessed datasets featuring front-facing images of the person and the product, often set against a clean background. Consequently, performance deteriorates when faced with unstructured real-world conditions, such as cluttered backgrounds, people wearing garment or less organized garments in reference images (shown in~\cref{fig:intro}c and~\cref{fig:intro}e). Robustness across diverse, in-the-wild scenarios is critical for wider adoption and usability. \textbf{3. Simple and Efficient Sampling Process}. Integrating reference images and pose conditions requires external image encoders that embed image conditions into the generation space. ControlNet ~\cite{zhang2023adding} based models ~\cite{kim2024stableviton, choi2024improving} are commonly used to facilitate the fusion of these images within the UNet framework. However, these additional modules increase computational overhead during both training and inference, impacting model efficiency and scalability. In addition, existing models rely on DDIM \cite{song2020denoising} for image sampling, which generally requires longer time steps.

In addition to these challenges, we believe that the scope of VTO should evolve beyond basic ``garment swapping'' to offer a more versatile, personalized virtual try-on experience across various product categories. This brings the following additional challenges: \textbf{4. Image Editing and Customization Capabilities}. Existing VTO models display limited flexibility in image editing, restricting their adaptability for practical applications. Essential functionalities—such as zero-shot generalization to previously unseen reference images, precise localized region editing and refinement, texture transfer, and object-level customization—are either only partially supported, absent, or require distinct models for each function (shown in~\cref{fig:intro}b and \cref{fig:intro}d). These capabilities are essential in real-world VTO production environments where users expect tailored adjustments and high degrees of personalization, making them vital for enhancing the practical utility and user satisfaction in VTO applications. \textbf{5. A Universal Model for Diverse VTO Product Types}. SOTA VTO models primarily focus on apparel, neglecting the diverse range of products customers may wish to virtually ``try on,'' including shoes, furniture, and accessories, etc (shown in~\cref{fig:intro}b and ~\cref{fig:intro}f). This often necessitates the development of separate models for different product types, reducing generality and underscoring the need for a unified Virtual Try-All (VTA) solution that can cater to multiple product categories seamlessly.

In this work, we present DiT-VTON, a novel approach aimed at tackling above key challenges in the VTO domain. By leveraging the Diffusion Transformer (DiT)-based text-to-image (T2I) architecture, renowned for its high-fidelity, realistic image generation, we adapt this framework specifically to the image-conditioned VTO task. Our approach begins with a structured exploration of various model configurations to achieve optimal integration of reference and masked images. These configurations include: 1) In-Context token concatenation: treating the reference image and masked image (source image with overlaid mask) as additional tokens in the input token sequence. 2) Channel concatenation: concatenating the noised image, masked image, reference image and mask along the channel dimension. 3) Control network integration: using a separate control network to fuse conditions into the main network's generation process. Additionally, we examine the enhancement of the generation process through pose conditions, achieved by stitching and concatenating pose information.  Our evaluations reveal that token concatenation with pose stitching yields the best performance. To further boost robustness and versatility, we train the optimal configuration on an expanded dataset featuring in-the-wild examples, such as images with cluttered backgrounds, unstructured reference images, and product categories beyond garments. Experimental results indicate that scaling the dataset in this manner significantly enhances model's performance on both wearable and non-wearable categories.

This paper makes four key contributions: 1) We examine various configurations of the Diffusion Transformer (DiT) to identify the optimal setup among them to take image conditions, enabling high-fidelity and efficient virtual try-on  2) We investigate different methods for incorporating pose control into the generation process, which further refines the model's precision. 3) We demonstrate that scaling the training data with beyond-wearable categories, enhances VTO performance and enables the model to support virtual try-on across diverse product categories (virtual try-all). 4) Beyond basic try-on functionality, our model also supports image editing, including fine-grained adjustments and object-level customization, increasing its practical applications.

  \section{Related Work}
\label{sec:related work}

\textbf{Image Generation for Virtual Try-On}
Virtual try-on~\cite{qi2024posevton,xu2025deft,han2024instructvton} refers to task that creates the image where a person naturally wears the provided garment. Early efforts focus on Generative Adversarial Network (GAN)~\cite{goodfellow2020generative} with a two-stage generation. First, a warping model deforms the given garment into a shape that approximately fits the person’s pose. Then, a GAN-based generator fuses the wrapped clothing into the result image generation. The training focuses more on the pixel-level alignment and how to more effectively fuse the wrapping module into the main generator~\cite{choi2021viton,ge2021parser,lee2022high,xie2023gp}. With the advent of more powerful Latent Diffusion Models~\cite{rombach2022high, esser2024scaling, nichol2021glide, saharia2022photorealistic},  attention has shifted to how to effectively integrate the garment details into the person’s context, including dual U-Net~\cite{zhu2023tryondiffusion}, encouraging sparse attention over the clothing area~\cite{kim2024stableviton}, texture and human identity preservation~\cite{yang2024texture, choi2024improving} and efficient training and inference~\cite{chong2024catvton}. In addition, VTO task is extended to multi-garment~\cite{li2024anyfit}, multi-view~\cite{wang2024mv}, multi-modal~\cite{zhang2024mmtryon}, style controlled~\cite{zhu2024m, chen2024wear}, multi-category virtual try-all~\cite{qi2024posevton,han2024instructvton,xu2025deft,seyfioglu2024diffuse,seyfioglu2023dreampaint}, mask-free try-on ~\cite{zhang2024boow}, etc.

\textbf{Image Composition}
Virtual try-on can be considered as a special subset of image composition. Image composition uplifts the garment constraint in the reference image and the mask area in the source image to be clothing. It aims to detect the “transferable” part in the reference image and naturally blend it into the mask area. Built upon the text based image editing diffusion models, Paint by example~\cite{yang2023paint} and Object stitch~\cite{song2023objectstitch} approach this task by using a CLIP image encoder to extract the representation of the object. ControlNet-based model structure is further employed in Anydoor~\cite{chen2024anydoor} to enhance the detail of the referenced objects. MimicBrush~\cite{chen2024zero} uses Dual U-Nets and video samples to achieve non-object image customization.  

\textbf{Diffusion Transformer for Image Generation}
Diffusion Transformers replace the U-Net with transformer blocks and are proven to generate high-quality images~\cite{peebles2023scalable}. FLUX models combine diffusion transformers with rectified flow achieve the state-of-the-art text to image generation and image editing abilities, meanwhile can significantly reduce the sampling steps. Though powerful, adapting Diffusion Transformers into inpainting tasks are rare. VITON-DiT~\cite{zheng2024vitondit} adapts diffusion transformers for video virtual try-on tasks. DiT \cite{alimama} provides a concurrent work that trains FLUX for text-based inpainting tasks in a ControlNet style. M\&M VTO~\cite{zhu2024m} combines the U-Net and DiT for garment and person encoding separately.


  \begin{figure*}[t]
  \centering
  \includegraphics[width=0.9\textwidth]{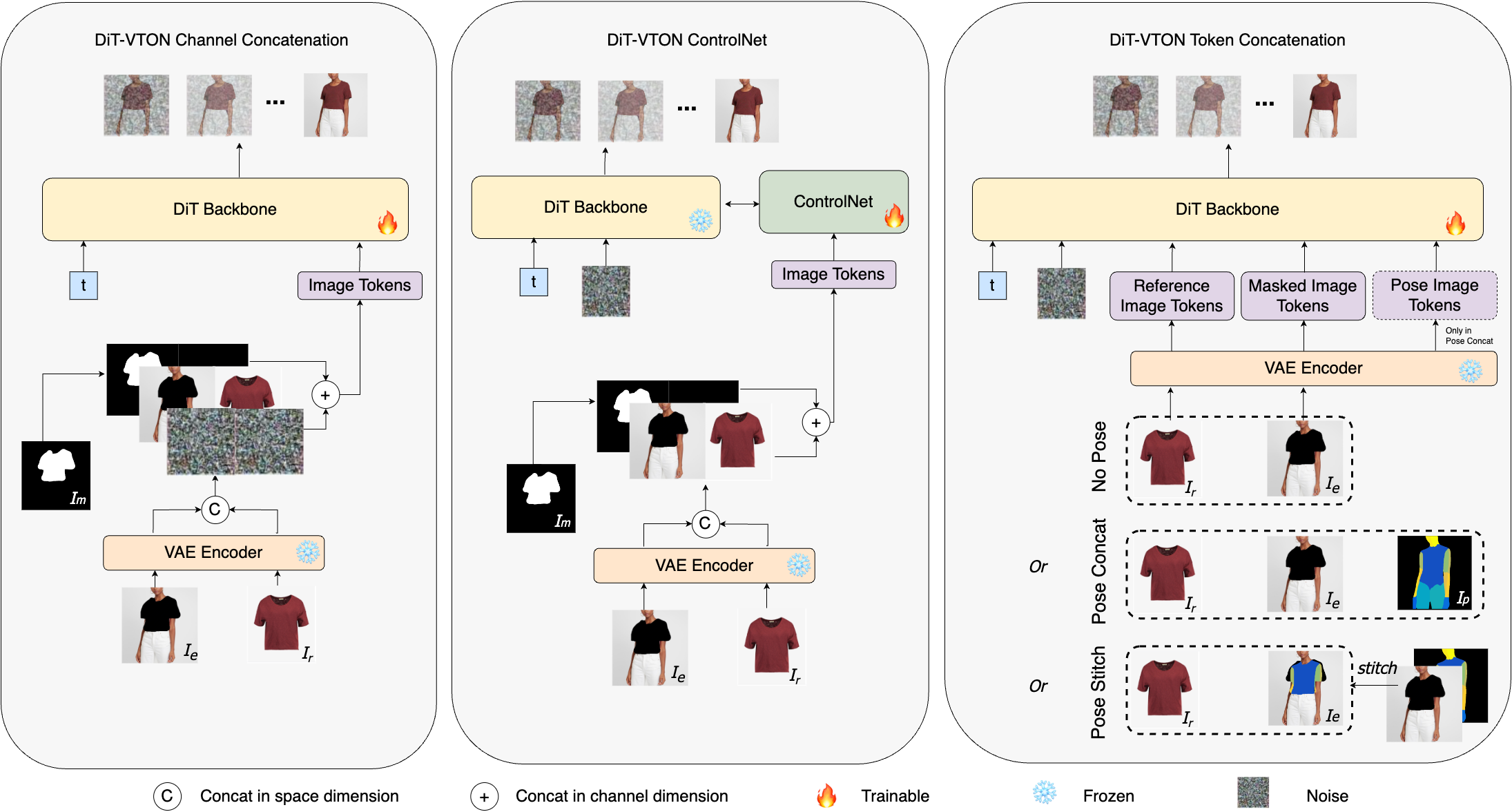}
  \caption{\bf{Illustration of different model configurations of DiT-VTON to effectively integrate image conditions.}}
  \label{fig:model_compare}
\end{figure*}

\section{Method}
\label{sec:method}

\subsection{Background}
\subsubsection{Diffusion Transformer}
Diffusion Transformers (DiTs) \cite{peebles2023scalable} introduce a new class of diffusion models that leverage transformer architecture for image generation. Diverging from traditional U-Net based approaches \cite{rombach2022high, podell2023sdxl}, DiTs replaced U-Net module with vision transformer blocks. As demonstrated in \cite{esser2024scaling}, DiT architecture achieves both higher efficiency and higher quality sample generation.





\subsubsection{Rectified Flow}

Rectified Flow \cite{liu2022flow} aim to simplify the denoising process in Diffusion models by creating smooth, continuous transformations from noise to data in a more direct manner. By rectifying or adjusting the flow of noise toward data points, these models can potentially achieve faster and more efficient image synthesis. Mathematically, rectified flow models leverage differential equations to trace paths directly through high-dimensional data space. Let \( x_t \) represent the sample at time \( t \) within the continuous flow. The rectified field \( \mathbf{f}_\theta(x_t, t) \) is trained to satisfy:

\[
\frac{d x_t}{dt} = \mathbf{f}_\theta(x_t, t),
\]
where \( \mathbf{f}_\theta \) learns a trajectory that smoothly interpolates between Gaussian noise and the data distribution. 




\subsection{Diffusion transformer for Virtual Try-On and Virtual Try-All (DiT-VTON)}
We formulate the VTO task as an image-conditioned inpainting problem. Given a source image $I_{s}$ of a person wearing garment A, a reference image $I_r$ of garment B, and a mask image $I_m$ specifying the area to be edited resulting in $I_e = I_s \odot I_m$, the model’s objective is to generate a target image $I_t$ where garment A in $I_{s}$ is replaced with garment B from $I_r$ . Each image input is first projected onto the latent space  $x = E(I)$ where $I \in \{I_s, I_m, I_r, I_e\}$, via an encoder $E$. Then, at each inference time step $t$, all the conditions along with the noise $x_t$ are integrated together as a combined input to the denoising transformer backbone $x_{in} = f(x_t, x_r, x_m, x_e)$, where $f$ represents different configurations to intake various image conditions by DiT.

\subsubsection{Image Encoder}
A typical choice in state-of-the-art VTO implementations is to use a dedicated garment encoder, which better preserves image features but also increases the number of model parameters, leading to higher computational costs and latency. As shown in the experimental results in subsequent sections, this added complexity is unnecessary; using a VAE encoder~\cite{pinheiro2021variational} directly can achieve comparable performance without any degradation in quality.

\begin{table*}[t]
\footnotesize
\centering
\caption{\textbf{Results of DiT-VTON's different model configurations to integrate image conditions.} (\cref{sec:model configuration}).}\label{tab:res model config}
\begin{tabular}{lrrrrlllrr}
\toprule
\multicolumn{1}{c}{\multirow{2}{*}{Model configuration}} & \multicolumn{4}{c}{VITON-HD} & \multirow{5}{*}{} & \multicolumn{4}{c}{DressCode} \\ \cmidrule(lr){2-5}\cmidrule(lr){6-10}
\multicolumn{1}{c}{} & \multicolumn{1}{l}{SSIM $\uparrow$} & \multicolumn{1}{l}{LPIPS $\downarrow$} & \multicolumn{1}{l}{FID $\downarrow$} & \multicolumn{1}{l}{KID $\downarrow$} &  & \multicolumn{1}{l}{SSIM $\uparrow$}  & \multicolumn{1}{l}{LPIPS $\downarrow$} & \multicolumn{1}{l}{FID $\downarrow$} & \multicolumn{1}{l}{KID $\downarrow$} \\ \midrule
DiT-VTON Channel Concat \emph{\scriptsize (Ours)} & \underline{0.9065} & \underline{0.0692} & \underline{9.263} & \underline{1.208} &  & \underline{0.9250} & \underline{0.0500} & \underline{8.345} & \underline{3.468} \\
DiT-VTON ControlNet \emph{\scriptsize (Ours)} & 0.8930 & 0.0861 & 10.734 & 2.202 &  & 0.9242 & 0.0639 & 9.804 & 4.721 \\
DiT-VTON Token Concat \emph{\scriptsize (Ours)} & \textbf{0.9130} & \textbf{0.0672} & \textbf{8.869} & \textbf{1.024} &  &  \textbf{0.9379} & \textbf{0.0451} & \textbf{5.498} & \textbf{1.349} \\ \bottomrule
\end{tabular}
\end{table*}

\begin{table*}[h]
\footnotesize
\centering
\caption{\textbf{Results of pose conditioning on best model configuration from Table~\ref{tab:res model config}, and comparison with SOTA VTO models.}}\label{tab:res baseline}
\begin{tabular}{lcccccccc}
\toprule
\multicolumn{1}{c}{\multirow{2}{*}{Models}} & \multicolumn{4}{c}{VITON-HD} & \multicolumn{4}{c}{DressCode} \\ \cmidrule(lr){2-5} \cmidrule(lr){6-9} 
\multicolumn{1}{c}{} & \multicolumn{1}{l}{SSIM $\uparrow$} & \multicolumn{1}{l}{LPIPS $\downarrow$} & \multicolumn{1}{l}{FID $\downarrow$} & \multicolumn{1}{l}{KID $\downarrow$} & \multicolumn{1}{l}{SSIM $\uparrow$} & \multicolumn{1}{l}{LPIPS $\downarrow$} & \multicolumn{1}{l}{FID $\downarrow$} & \multicolumn{1}{l}{KID $\downarrow$} \\ \midrule
StableVTON~\cite{kim2024stableviton} & 0.8543 & 0.0905 & 11.054 & 3.914 & - & - & - & - \\
LaDI-VTON~\cite{morelli2023ladi} & 0.8603 & 0.0733 & 14.648 & 8.754 & 0.7656 & 0.2366 & 10.676 & 5.787 \\
IDM-VTON~\cite{choi2024improving} & 0.8499 & 0.0603 & 9.842 & 1.123 & 0.8797 & 0.0563 & 9.546 & 4.320 \\
OOTDiffusion~\cite{xu2024ootdiffusion} & 0.8187 & 0.0876 & 12.408 & 4.680& 0.8854 & 0.0533 & 12.567 & 6.627 \\
CatVTON~\cite{chong2024catvton} & 0.8704 & \underline{0.0565} & 9.015 & 1.091 & 0.8922 & 0.0455 & 6.137 & \underline{1.403} \\ \midrule
DiT-VTON Token Concat \emph{\scriptsize (Ours)} & 0.9130 & 0.0672 &  8.869 & 1.024 & 0.9379 & 0.0451 & \textbf{5.498} & \textbf{1.349} \\
DiT-VTON Token Concat Pose Concat \emph{\scriptsize (Ours)} & \textbf{0.9229} & \textbf{0.0560} & \textbf{8.628} & \underline{0.881} & \underline{0.9430} & \underline{0.0397} & 6.149 & 2.028 \\
DiT-VTON Token Concat Pose Stitch \emph{\scriptsize (Ours)} & \underline{0.9216} & 0.0576 & \underline{8.673} & \textbf{0.820} & \textbf{0.9432} & \textbf{0.0389} & \underline{5.584} & 1.540 \\ 
\bottomrule
\end{tabular}
\end{table*}

\subsubsection{Image Condition Integration}
\label{sec:model configuration}
We explore the optimal model configuration ($f$) to integrate additional image conditions into the transformer blocks. The architectures are shown in~\cref{fig:model_compare}.
\begin{itemize}

\item \textbf{Channel Concatenation} (``Channel Concat''). We follow conventional approach in U-Net-based inpainting models~\cite{Rombach_2022_CVPR} that concatenates the masked image $I_e$, the mask image $I_m$ and the latent noise $x_t$ in the channel dimension. As for the additional reference image, we concatenate it with the masked image at the spatial dimension, as similar practice in~\cite{choi2024improving, yang2024texture, chong2024catvton}: $x_{c} = x_e\oplus x_r$, $m_c = x_m\oplus \mathcal{O}$, and $z_{c} = x_t\oplus \mathcal{P}$, where $\oplus$ denotes spatial concatenation and $\mathcal{O}$ and $\mathcal{P}$ is the padding image. The final input to the transformer is $P(x_{c}\copyright m_c \copyright z_{c})$, where $\copyright$ is channel concatenation. To get the output, we only keep the first half of the output in the spatial dimension.

\item \textbf{ControlNet}. Another conventional way of adding control to diffusion models is by copying (part of) the main denoising backbone as the ControlNet to encode conditions. Then the encoded image representation is fused back to the main backbone by cross attention, summation or other adaptive norm layers~\cite{peebles2023scalable, karras2019style}. Here, we concatenate $x_{c}$ and $m_c$ in the channel dimension as the input to ControlNet. $z_c$ is the input to the main transformer blocks.

\item \textbf{Token Concatenation} (``Token Concat''). Inspired by previous work~\cite{peebles2023scalable, flux2024, zhou2024transfusion}, we patchify (denoted as $P$) each latent image into tokens and directly concatenate all the image tokens together as input: $P(x_t)\circ P(x_r)\circ P(x_e)$, where $\circ$ denotes concatenation in sequential dimension. After the final block, we remove the conditioning tokens from the sequence. Here, we drop the latent mask. 
\end{itemize} 
\textbf{Adding pose constraint}.
In a concurrent work \cite{qi2024posevton}, we investigate methods to preserve pose information. In alignment with that work, here we explore two approaches to recover pose details hidden by the mask, by conditioning on a pose image $I_p$: 1) \textbf{Pose Concatenation} (``Pose Concat''): Adding the pose image tokens to the input sequence. While simple, it results in longer input with additional computation and inference latency. 2) \textbf{Pose Stitching} (``Pose Stitch''): Stitching the pose image into the masked area $I_e = I_s \odot I_m + I_p \odot (1 - I_m) $. Both methods are applied to the best performing approach, see Figure~\ref{fig:model_compare} right (Token Concatenation architecture) for an illustration of the two methods (bottom two rows of inputs). For more comprehensive experiments and analysis of pose conditioning strategies, we refer the reader to \cite{qi2024posevton}.

  \begin{figure*}[t]
  \centering
  \includegraphics[width=\textwidth]{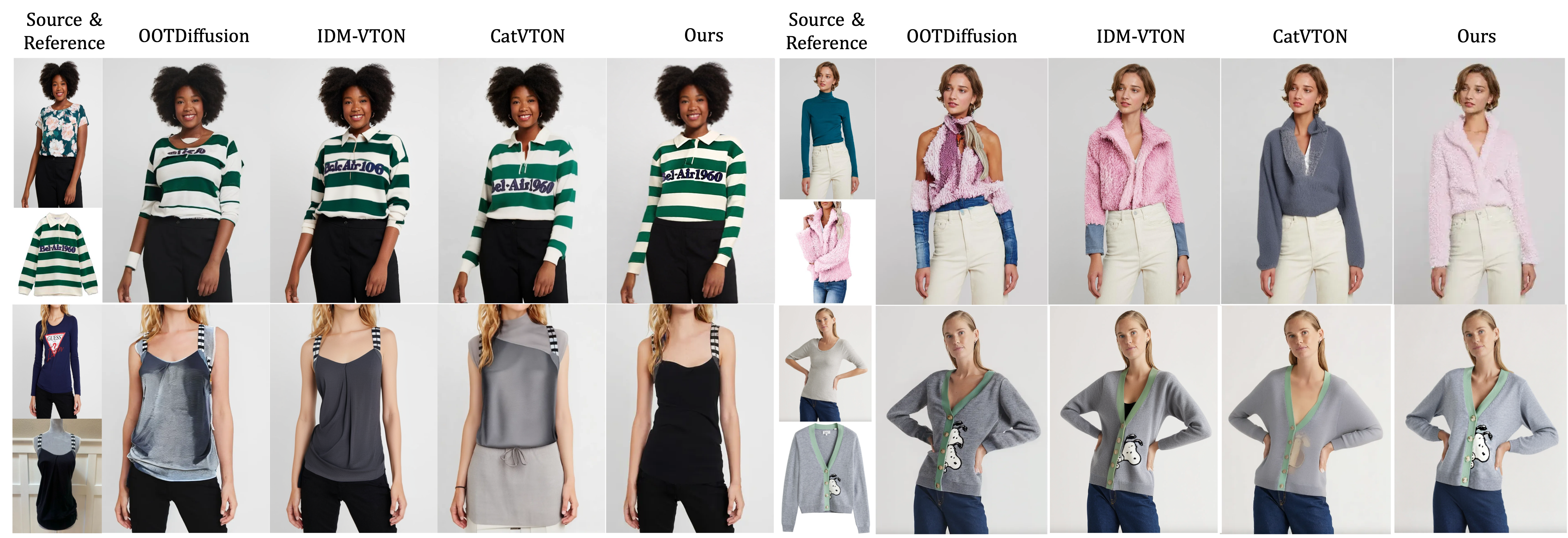}
  \caption{\textbf{Comparison of DiT-VTON with state-of-the-art models on virtual try-on test cases.} Our method can generate better garment details and layouts, \underline{e.g. number of buttons and puppy's eyes/paws in bottom right example}. Please zoom in to see details preservation.}
  \label{fig:qual vto}
\end{figure*}
\begin{figure*}[t]
  \centering
  \includegraphics[width=0.95\textwidth]{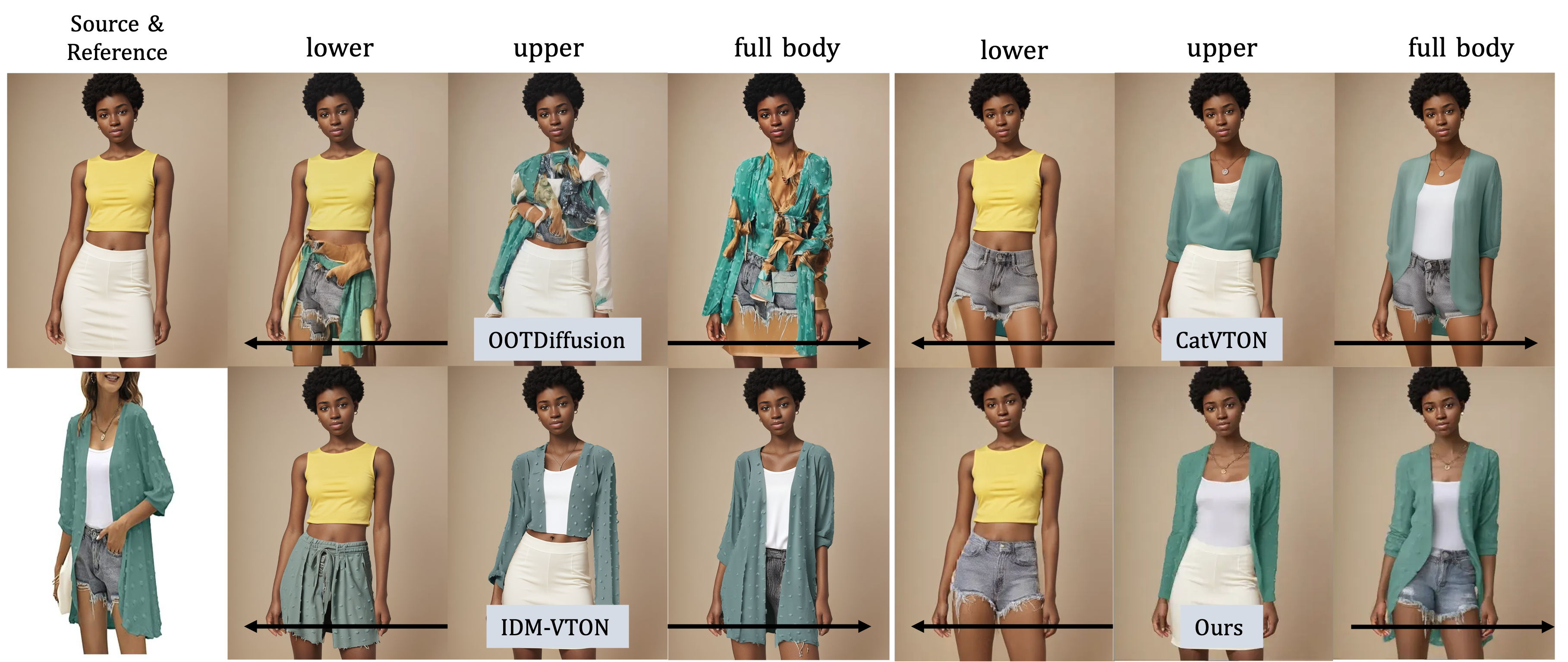}
  \caption{\textbf{Comparison of DiT-VTON with state-of-the-art models on multi-garment try-on.} Our method can generate better garment details and layouts. Please zoom in to see details preservation.}
  \label{fig:qual set}
\end{figure*}

\section{Experiments}
\label{sec:experiments}

\subsection{Datasets}
\textbf{Virtual Try-on dataset} To evaluate the best model configuration for integrating image conditions in the VTO task, we conduct experiments using the test sets of two widely recognized public benchmarks: VITON-HD~\cite{choi2021viton} and DressCode~\cite{morelli2022dress}. Each data pair in these datasets consists of a reference product image and a source image, where the source image serves as the target label, depicting the model wearing the product from the reference image. This setup allows the model to learn how to realistically overlay the product onto the source image.VITON-HD includes only upper garments in front-facing poses, with a total of 2,032 pairs for testing. For our experiments, we utilize the preprocessed mask and pose images provided with the dataset. DressCode offers a more diverse set of 5,400 test pairs, covering upper and lower garments as well as full-body dresses. For consistency in pose representation, we apply the same colormap as used in VITON-HD. To obtain clothing-agnostic masks, we use auto-masker prediction scripts. To alleviate the mask overfitting issue, we randomly sample 50\% of the training data and use the minimal bounding box that covers the clothing-agnostic masks.

\textbf{Virtual Try-all dataset} Beyond the limitations of front-facing garment try-on and commonly used garment categories like shirts and pants, we collected a new dataset designed to cover more diverse clothing and product categories, as well as to handle more complex, in-the-wild scenarios. This dataset includes over 1,000 different object categories and is structured as follows: 1) Expanded Clothing Categories: Examples include jackets, pajamas, and suits, etc; 2) Wearable and Non-Wearable Items: Wearable items include bags, hats, and shoes, while non-wearable items encompass jewelry, devices, toys, and more. 3) In-the-Wild Source and Reference Images: Source images may contain individuals in various background scenes, with subjects not necessarily facing forward. Reference images may not always present the product in isolation; for instance, clothing might appear in both front and back views or shown flat instead of modeled in 3D forms. 

\textbf{Data cleaning and preparation} To ensure accurate image pair selection, we first apply a classifier to verify whether the source image is displaying the same product shown in the reference image. For masks, we first apply a text-prompted object detection model to generate the bounding box area of the product in the source image using the product category as the text prompt. Then, we use the bounding box to prompt an object segmentation model to create the mask. 

\subsection{Implementation details}
We use a diffusion transformer architecture as our diffusion backbone, building upon the implementation provided in the Diffusers library~\footnote{https://github.com/huggingface/diffusers}. All images are padded and resized to $512 \times 512$ pixels. The model is trained with a learning rate of 1e-5 and a batch size of 32. During inference, we use a sampling step of 28. The DiT model is tuned on 4 NVIDIA H100 80GB GPUs for 5,000 iterations. 

\begin{figure*}[t]
  \centering
  \includegraphics[width=0.85\textwidth]{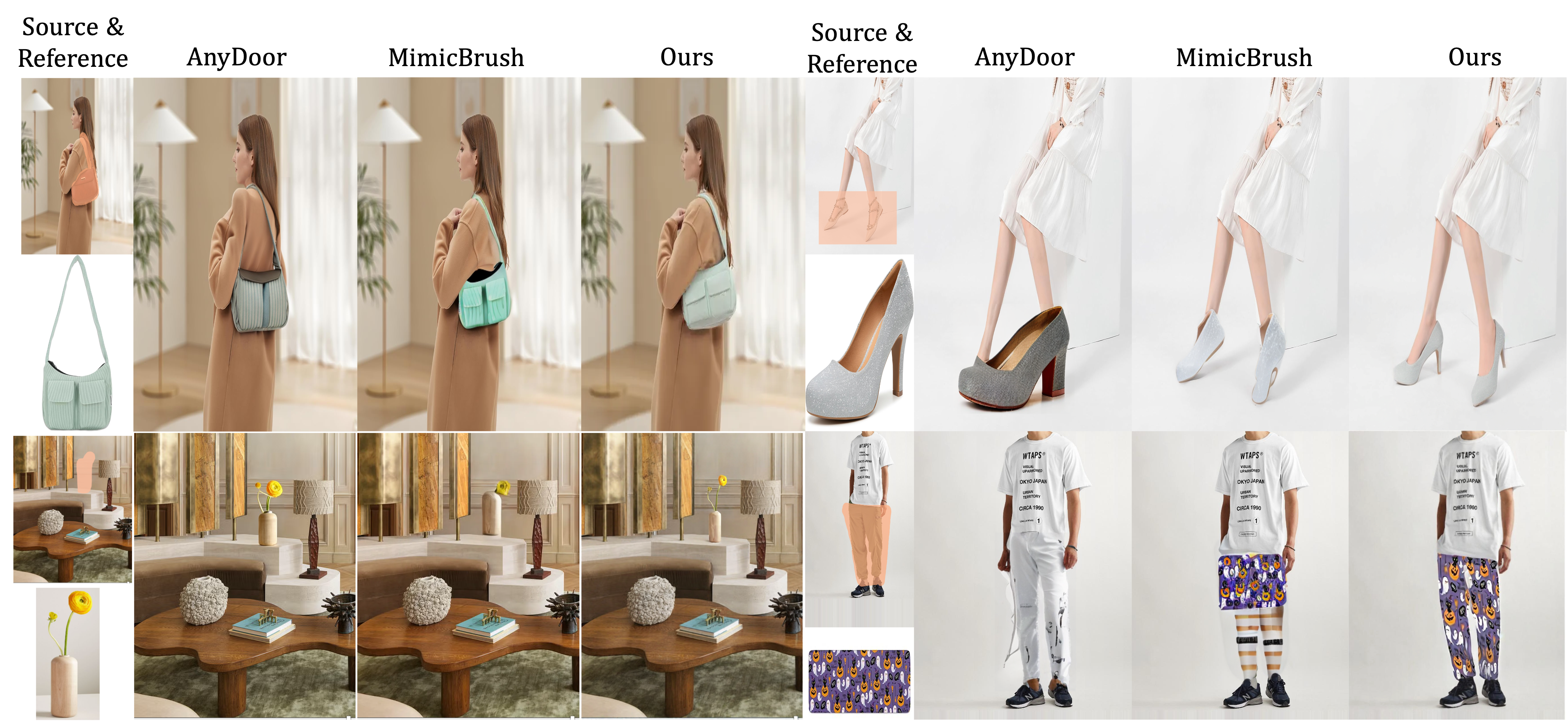}
  \caption{\textbf{Comparison of DiT-VTON with state-of-the-art models on generalization ability of virtual try-all (VTA)/image editing.} Our method can generate better garment details and layouts, \underline{e.g. the texture of vase in bottom left example}. Please zoom in to see details.}
  \label{fig:qual vta}
\end{figure*}
\subsection{Evaluation Metrics}
The evaluation is conducted in both paired and unpaired settings. In the paired setting, the garment in the source image matches the garment in the reference image, with the source image serving as the ground truth. We use Structural Similarity Index (SSIM)~\cite{wang2004image} and Learned Perceptual Image Patch Similarity (LPIPS)~\cite{zhang2018unreasonable} to compare the ground truth with the generated results. In the unpaired setting, the garment in the source image differs from that in the reference image. Here, we compute the distributional distance between the source and generated images using Fréchet Inception Distance (FID)~\cite{zhang2018unreasonable} and and Kernel Inception Distance (KID)~\cite{binkowski2018demystifying}. We follow the implementation in~\cite{morelli2023ladi}.

\subsection{Quantitative Results}
\label{sec:quant}
In \cref{sec:quant_best_model} - \cref{sec:pose_quant}, we quantitatively demonstrate that token concatenation with pose stitching (both defined in \cref{sec:model configuration}) yield our best-performing model configuration. Additionally, we evaluate its generalization across diverse product categories in \cref{sec:vta_quant}.
\subsubsection{Comparison between model configurations} 
\label{sec:quant_best_model}
We evaluate all 3 proposed conditional model configurations  (\cref{sec:model configuration}) to identify the optimal image integration setup for VTO task among the explored ones. \cref{tab:res model config} presents the evaluation results on the VITON-HD and DressCode datasets. Our findings indicate that using concatenated image tokens as model inputs yields the best performance. This approach is closely followed by channel concatenation, which outperforms the ControlNet-based structure. Notably, this performance hierarchy remains consistent across both the VITON-HD and DressCode datasets. 


\subsubsection{Comparison with baseline SOTA models} We compare our approach with state-of-the-art methods as outlined in~\cite{chong2024catvton}, a comprehensive and recent study in the field. As shown in~\cref{tab:res baseline}, our model, which employs image token concatenation, significantly outperforms the established VTO models across all evaluated metrics on both the VITON-HD and DressCode datasets. 

\subsubsection{Pose preservation}
\label{sec:pose_quant}
In addition, in~\cref{tab:res baseline}, we demonstrate the impact of adding pose conditioning to our models and compare two integration strategies. Overall, incorporating pose condition enhances model performance, although it introduces a slight decrease in unpaired generalization (as indicated by the KID metric) on the DressCode dataset. When comparing the two integration strategies—pose concatenation and pose stitching—we observe that their performances are closely comparable. Specifically, pose concatenation yields better results on the VITON-HD dataset, whereas pose stitching is more effective on the DressCode dataset. 

\begin{table}[h]
\setlength{\tabcolsep}{6pt} 
\footnotesize 
\centering
\caption{\textbf{Quantitative results of DiT-VTON compared with SOTA models on general product categories (Virtual Try-All).} AnyDoor/Mimicbrush[original] denotes open-source publicly available checkpoints, while AnyDoor/Mimicbrush[vitall] represents training the respective model on our Virtual Try-All dataset (as an ablation on both the model/method and the dataset).}
\label{tab:res_vitall}
\begin{tabular}{lcccc}
\toprule
\multirow{2}{*}{Model [Training Dataset]} & \multicolumn{2}{c}{Non-wearable} & \multicolumn{2}{c}{Wearable} \\ 
\cmidrule(lr){2-3} \cmidrule(lr){4-5} 
& SSIM $\uparrow$ & LPIPS $\downarrow$ & SSIM $\uparrow$ & LPIPS $\downarrow$ \\ 
\midrule
AnyDoor[original] \cite{chen2024anydoor}  & 0.7313 & 0.2954 & 0.7442 & 0.2787 \\
AnyDoor[vitall] & 0.7766 & 0.2324 & 0.7857 & 0.2157 \\ 
MimicBrush[original] \cite{chen2024zero} & 0.9060 & 0.0924 & 0.8852 & 0.1100 \\ 
MimicBrush[vitall] & 0.9078 & 0.0910 & 0.8998 & 0.0896 \\ 
\midrule
DiT-VTON[vton] \emph{(Ours)} & \underline{0.9088} & \underline{0.0898} & \underline{0.9082} & \underline{0.0731} \\ 
DiT-VTON[vitall] \emph{(Ours)} & \textbf{0.9281} & \textbf{0.0617} & \textbf{0.9171} & \textbf{0.0617} \\ 
\bottomrule
\end{tabular}
\end{table}


\subsubsection{Generalization beyond garment try-on}
\label{sec:vta_quant}
Beyond basic garment try-on, we evaluate the generalization capabilities of our best model configuration, token concatenation with pose stitching, using our in-house collected Virtual Try-All dataset. First, we assess the model in a zero-shot scenario, training it exclusively on garment categories (denoted as DiT-VTON[vton]) and comparing its performance against Anydoor~\cite{chen2024anydoor} and MimicBrush~\cite{chen2024zero} on both wearable and non-wearable test sets. As shown in the first section of ~\cref{tab:res_vitall}, DiT-VTON[vton] demonstrates notable zero-shot capabilities, even when trained solely on garment categories. Next, we extend the training set to include a broader range of product categories and re-evaluate the model on the Virtual Try-All dataset (refer to the last row in~\cref{tab:res_vitall}). This expanded model, denoted DiT-VTON[vitall] outperforms the VTON-specific model across both test sets. Additionally, we present results obtained via training AnyDoor/MimicBrush with our Virtual Try-All dataset  (denoted as AnyDoor/MimicBrush[vitall]) as an ablation of the dataset from the model. These results show that scaling and diversifying the training data helps with model performance across categories, both wearable and non-wearable. 


\begin{figure}[t]
  \centering
  \includegraphics[scale=0.35]{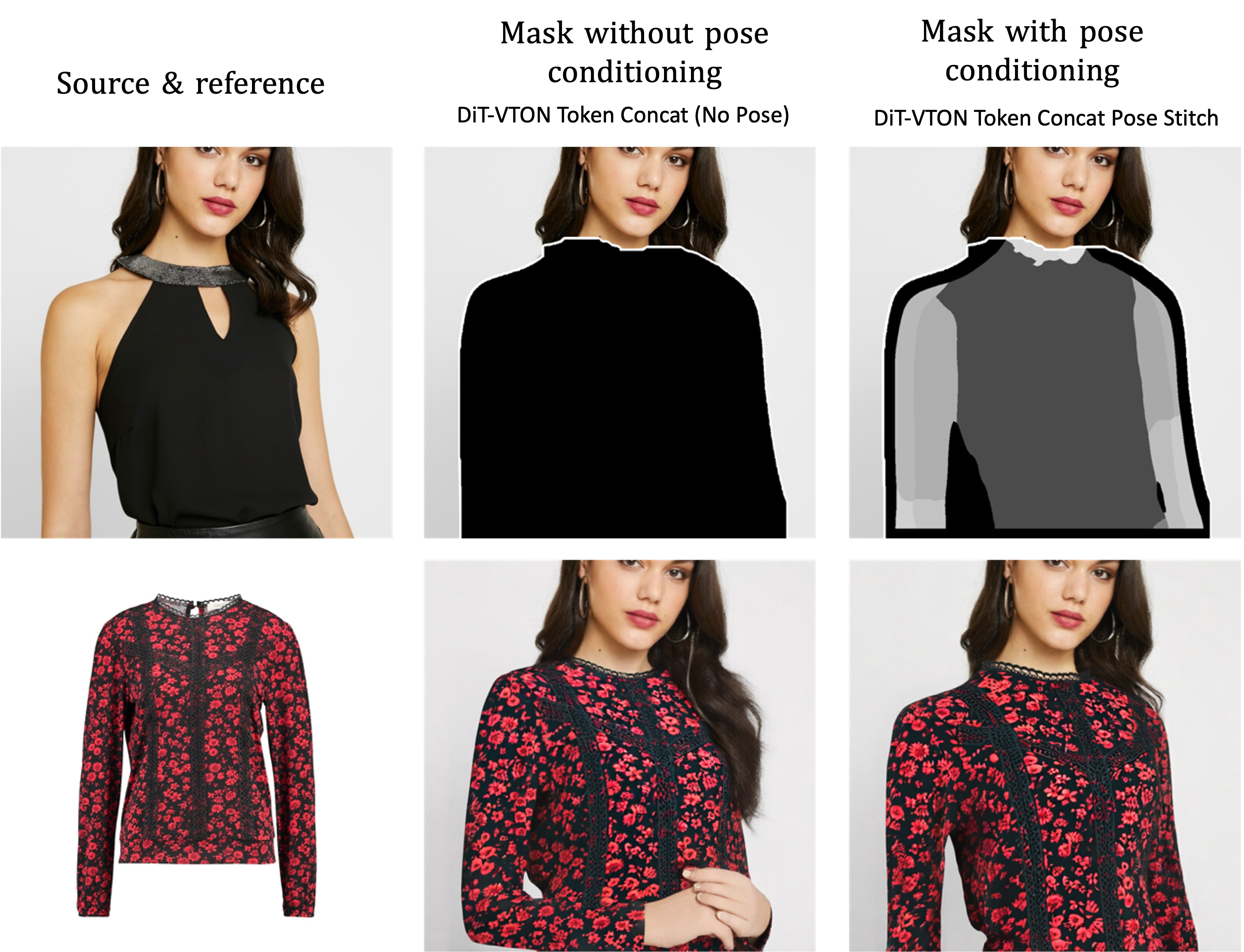}
  \caption{\textbf{Comparison of DiT-VTON model configuration with and without pose conditioning.} See more figures of pose preservation experiments in the Supplementary Material.}
  \label{fig:result}
\end{figure}


\subsection{Qualitative results}
In this section, we present qualitative examples to illustrate the superior robustness, generalization capabilities, versatility, and strong image customization abilities of our DiT-VTON model. All results are generated using our best-performing model configuration (namely DiT-VTON Token Concat Pose Stitch), as quantitatively evidenced in ~\cref{sec:quant}. 
\subsubsection{Comparison with baseline models}
Using in-the-wild test examples, we provide pairwise comparisons between the outputs of our method and those of baseline models in ~\cref{fig:qual vto}. As shown, our model consistently excels in preserving key attributes: texture in the striped shirt, styling in the pink coat, color in the black sleeveless shirt, and details in the cartoon character graphic. We further evaluate our model in a multi-garment try-on scenario. Using a full-body outfit reference image, our model accurately captures the clothing ``semantics'' from the reference and generates high-fidelity results for lower-body, upper-body, and full-body try-on.

\subsubsection{Generalization beyond garment try-on}
Our model’s performance also generalizes to categories beyond garments, such as hats, handbags, jewelry, furniture, etc. As illustrated in~\cref{fig:qual vta}, compared to Anydoor~\cite{chen2024anydoor} and MimicBrush~\cite{chen2024zero}, our model generates more photorealistic and style-preserved results on bags and shoes. Furthermore, as shown in the second row in~\cref{fig:qual vta}, our model outperforms the baseline models on local editing and texture transfer.

\subsubsection{Pose preservation}
In~\cref{fig:result}, we compare results generated by our DiT-VTON model from the same source image, with and without pose conditioning. This comparison illustrates that pose conditioning avoids hallucination of arm poses and hands and helps with the preservation of source pose, consistent with the quantitative findings.




\section{Conclusion and Limitations}
In this paper, we explored the use of Diffusion Transformer for the Virtual Try-on (VTO) task. We investigated three model configurations—Token Concatenation, Channel Concatenation, and ControlNet—for integrating image conditions, along with two pose preservation strategies: Pose Concatenation and Pose Stitching. Our experimental results demonstrate that Token Concatenation with Pose Stitching consistently outperforms state-of-the-art methods on two widely used public datasets, VITON-HD and DressCode. Additionally, we propose an innovative redefinition of VTO, broadening it beyond basic ``garment try-on'' to a more versatile and personalized task. Building on this concept, we present DiT-VTON, an all-purpose virtual try-on model that not only generalizes to diverse product categories for ``Virtual Try-all'' (VTA) applications, but is also integrated with multiple image customization abilities including pose preservation, precise localized region editing and refinement, texture transfer and object-level customization. Our model quantitatively and qualitatively outperforms state-of-the-art VTO models across all functionalities and demonstrates enhanced robustness when applied to in-the-wild and even low-resolution images. As a limitation, DiT-VTON may struggle with certain product categories, particularly those that occupy a small area within reference images. For these cases, the generated results tend to show blurry artifacts around the mask edges. 


  {
      \small
      \bibliographystyle{unsrt}
      \bibliography{main}
  }
\clearpage
\setcounter{page}{1}
\maketitlesupplementary

\section{More Qualitative Examples}
\begin{figure*}[h]
  \centering
  \includegraphics[width=0.9\textwidth]{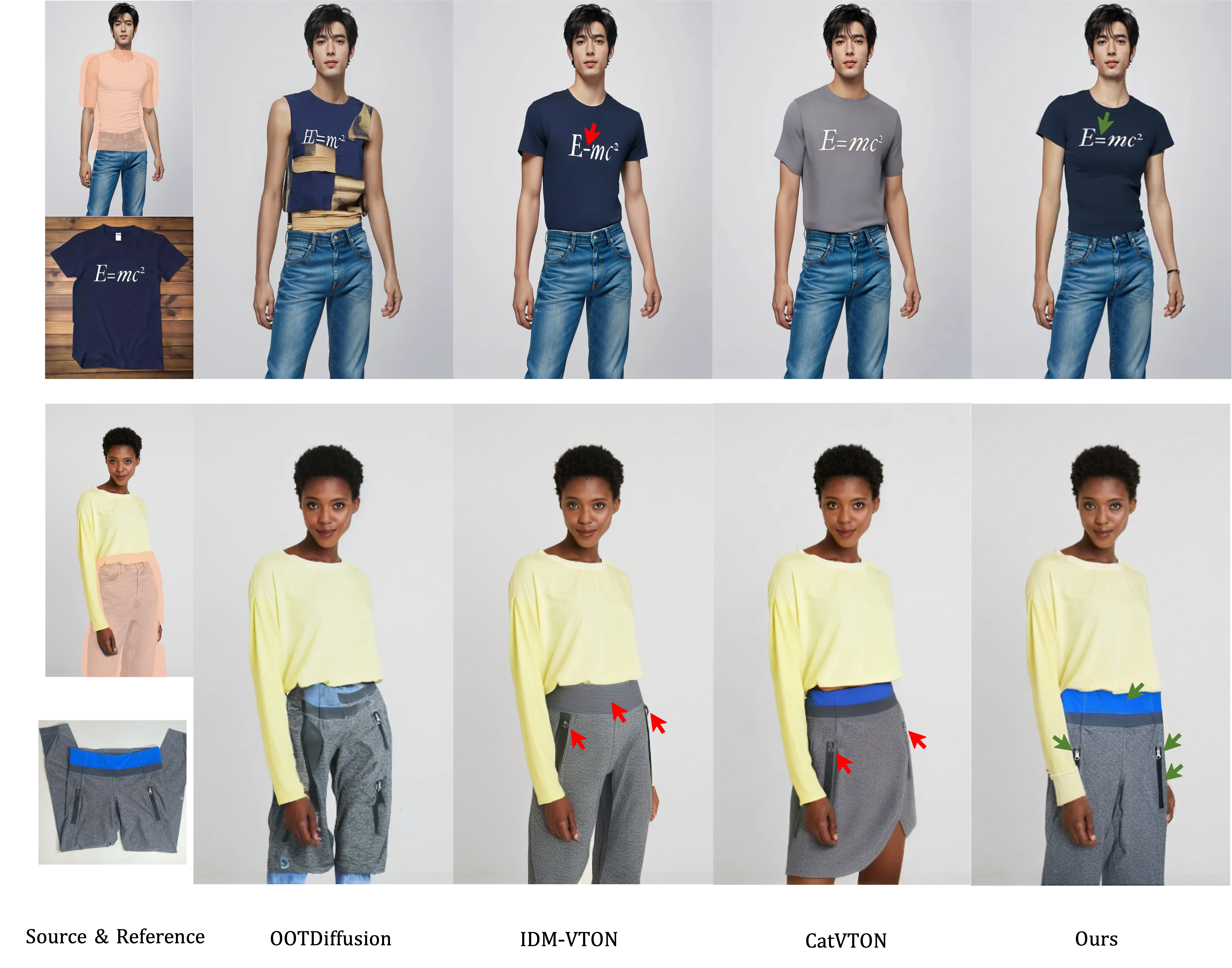}
  \caption{\textbf{In-the-wild try on: unorganized reference image and text preservation.} \textcolor{red}{Red} arrow highlights error of alternative state-of-the-art models. \textcolor{Green}{Green} arrow highlights correct output from our model. Our method can generate better garment details and layouts. Please zoom in to see details preservation.}
  \label{fig:qual1}
\end{figure*}

\begin{figure*}[h]
  \centering
  \includegraphics[width=0.9\textwidth]{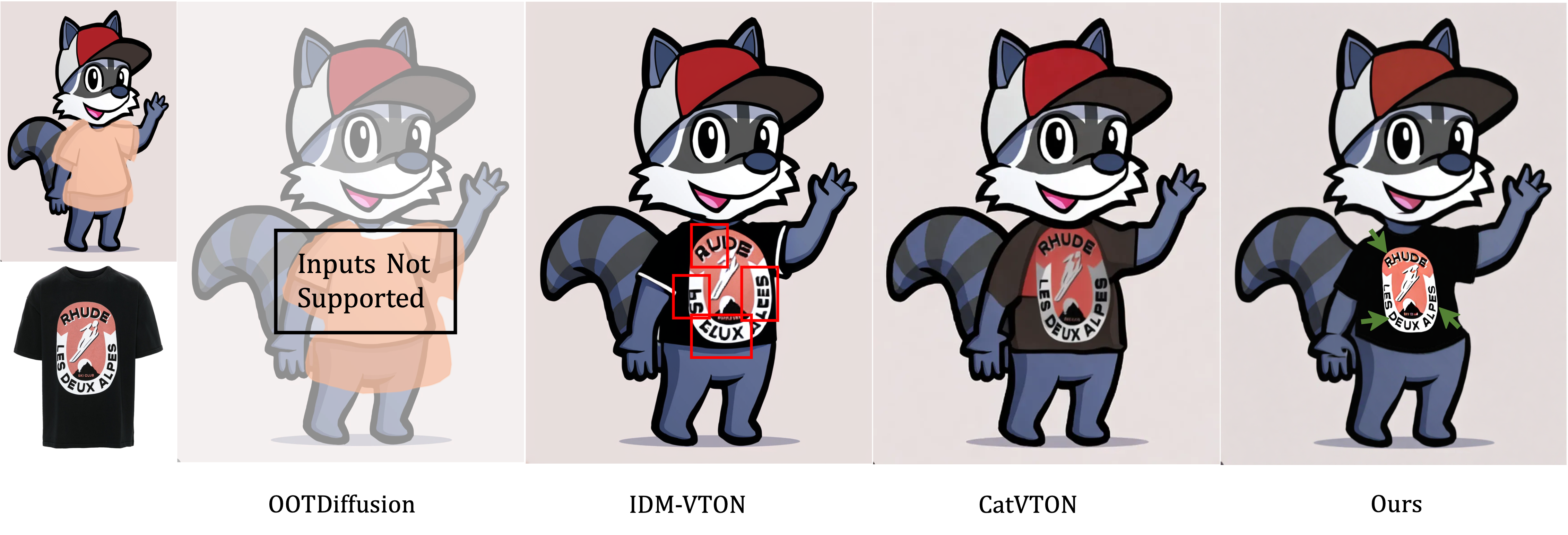}
  \caption{\textbf{In-the-wild try on: in-the-wild source image and text preservation.} \textcolor{red}{Red} boxes highlight errors of alternative state-of-the-art models. \textcolor{Green}{Green} arrow highlights correct output from our model. Our method can better preserve subtle garment details and layouts. Please zoom in to see details preservation.}
  \label{fig:qual3}
\end{figure*}

\begin{figure*}[h]
  \centering
  \includegraphics[width=0.9\textwidth]{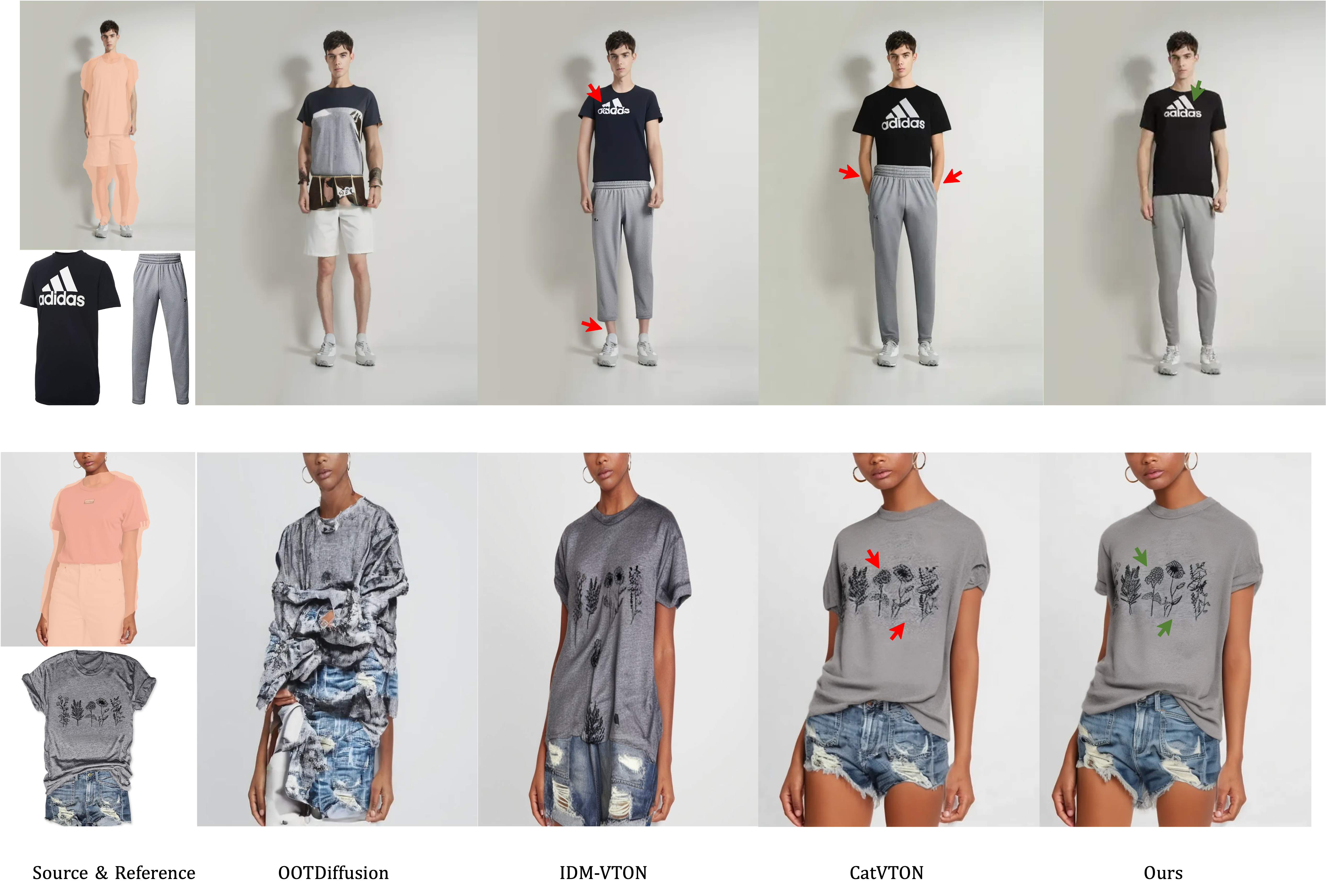}
  \caption{\textbf{Multi-garment try on.} \textcolor{red}{Red} arrow highlights less clarity details generated from alternative state-of-the-art models. \textcolor{Green}{Green} arrow highlights correct output from our model. Our method can generate better and higher clarity garment details and layouts. Please zoom in to see details.}
  \label{fig:qual4}
\end{figure*}

\begin{figure*}[h]
  \centering
  \includegraphics[width=0.9\textwidth]{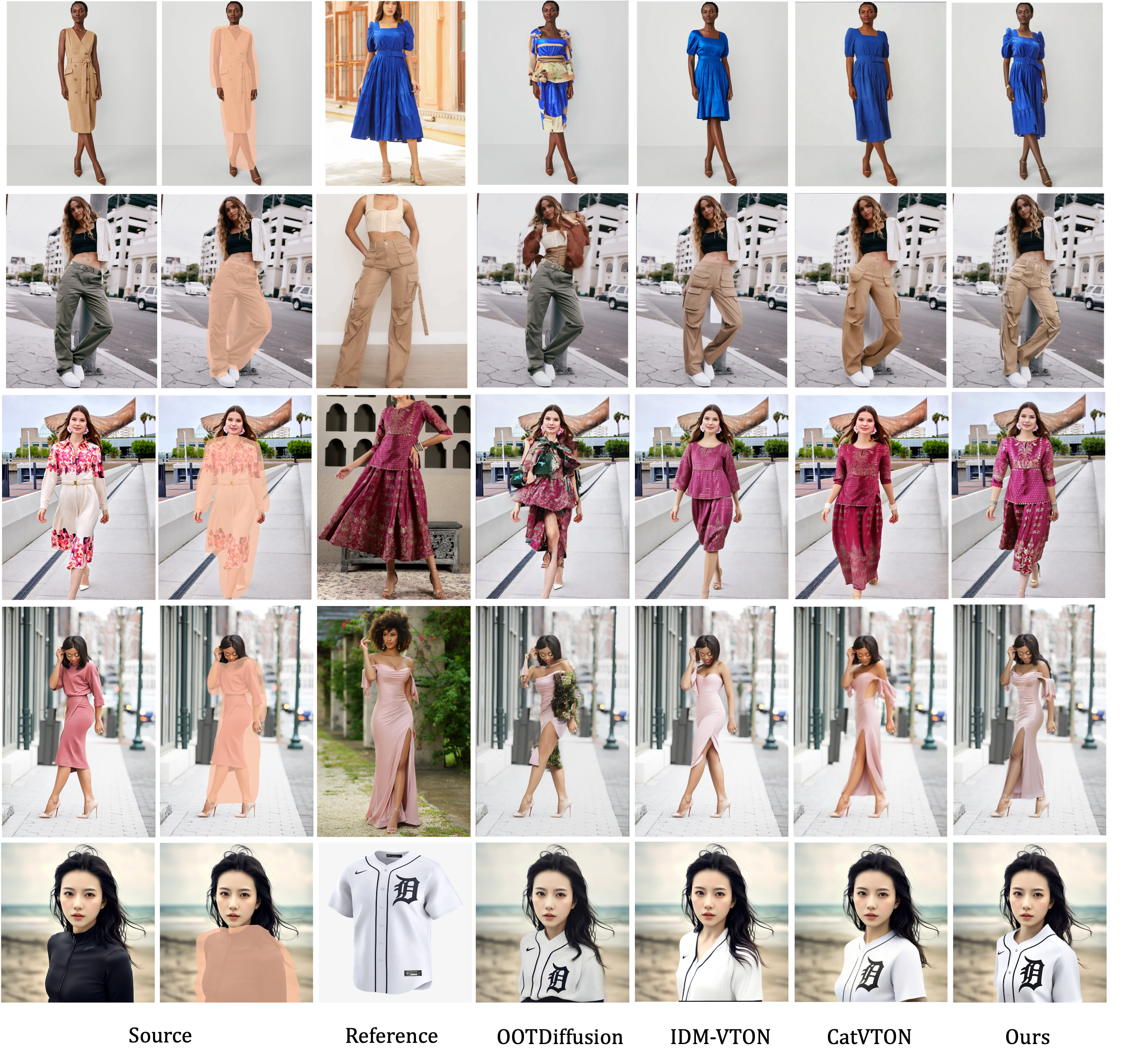}
  \caption{\textbf{In-the-wild try on: person-wearing reference image, with diverse background in source and reference images.} Our method can generate better garment details and layouts. Please zoom in to see details preservation.}
  \label{fig:more_vto}
\end{figure*}

\begin{figure*}[h]
  \centering
  \includegraphics[width=0.9\textwidth]{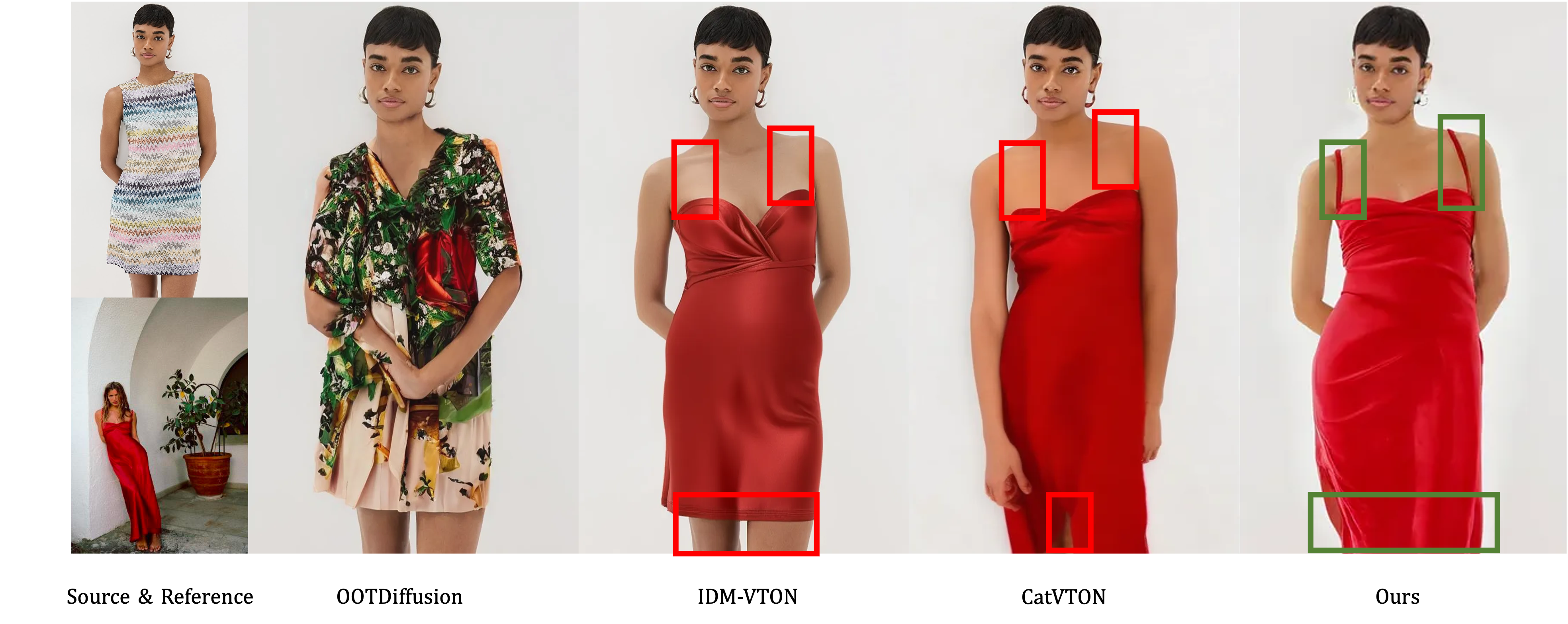}
  \caption{\textbf{In-the-wild try on: person-wearing reference image.} \textcolor{red}{Red} boxes highlight errors of alternative state-of-the-art models. \textcolor{Green}{Green} boxes highlight correct output from our model.}
  \label{fig:qual2}
\end{figure*}

\begin{figure*}[h]
  \centering
  \includegraphics[width=0.8\textwidth]{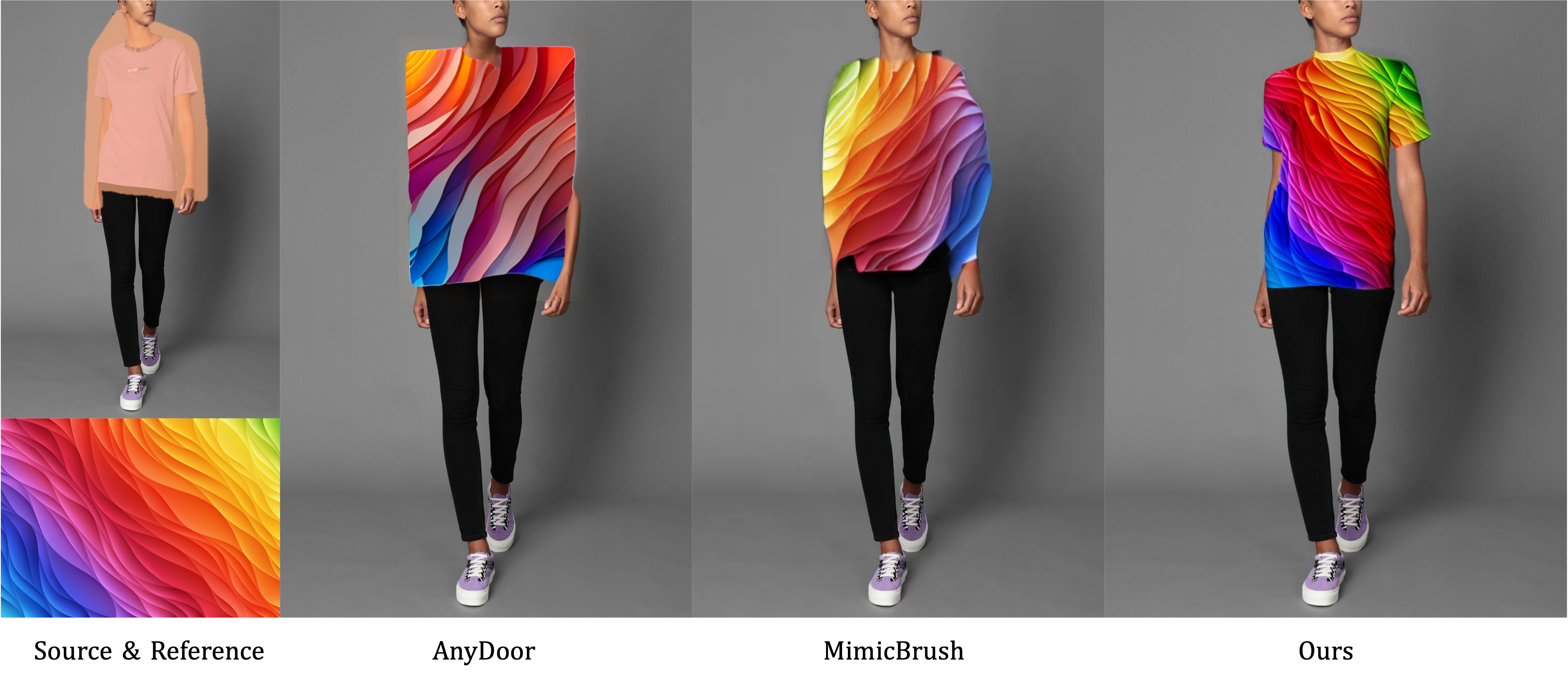}
  \caption{\textbf{Image customization: texture transfer.}}
  \label{fig:qual5}
\end{figure*}

\begin{figure*}[h]
  \centering
  \includegraphics[width=0.85\textwidth]{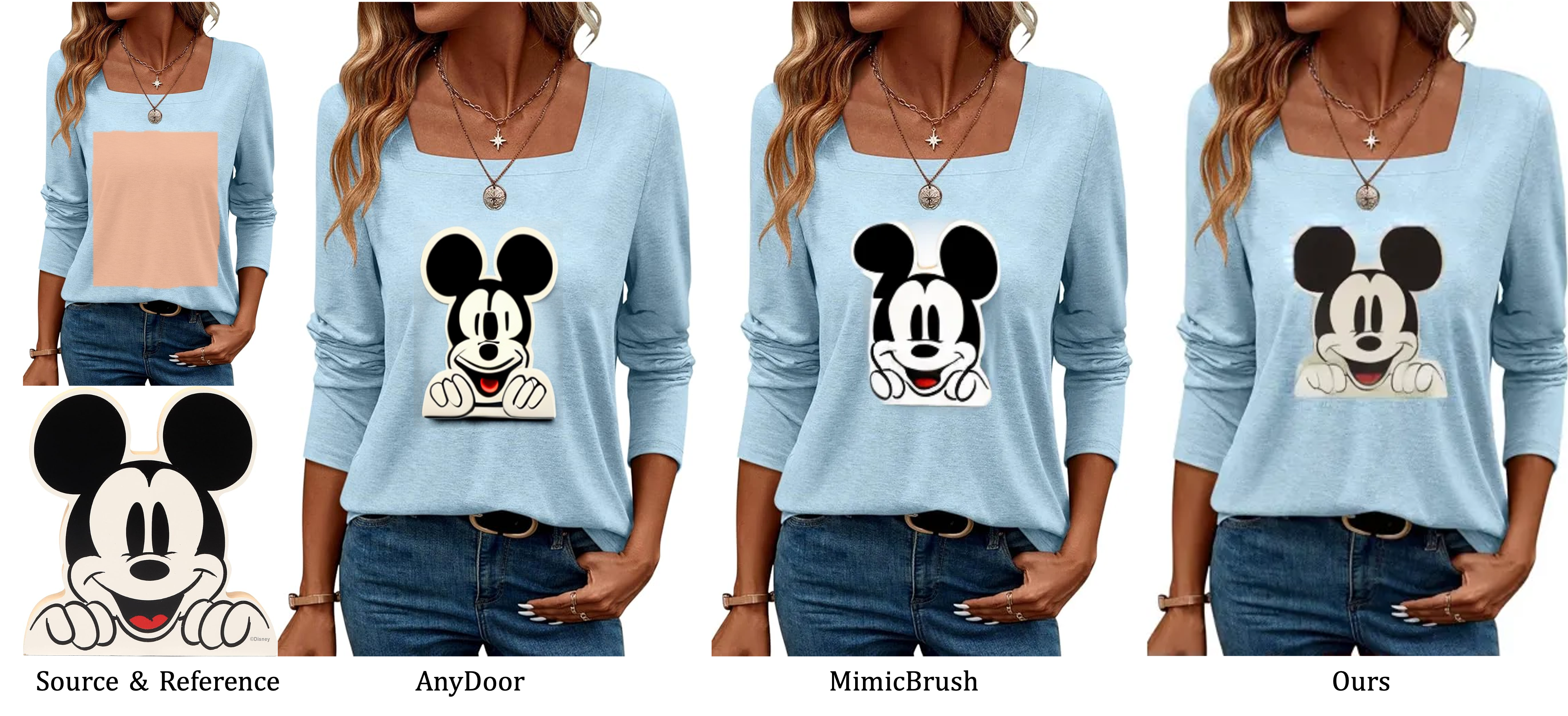}
  \caption{\textbf{Image customization: local editing.}}
  \label{fig:qual6}
\end{figure*}

\begin{figure*}[h]
  \centering
  \includegraphics[width=\textwidth]{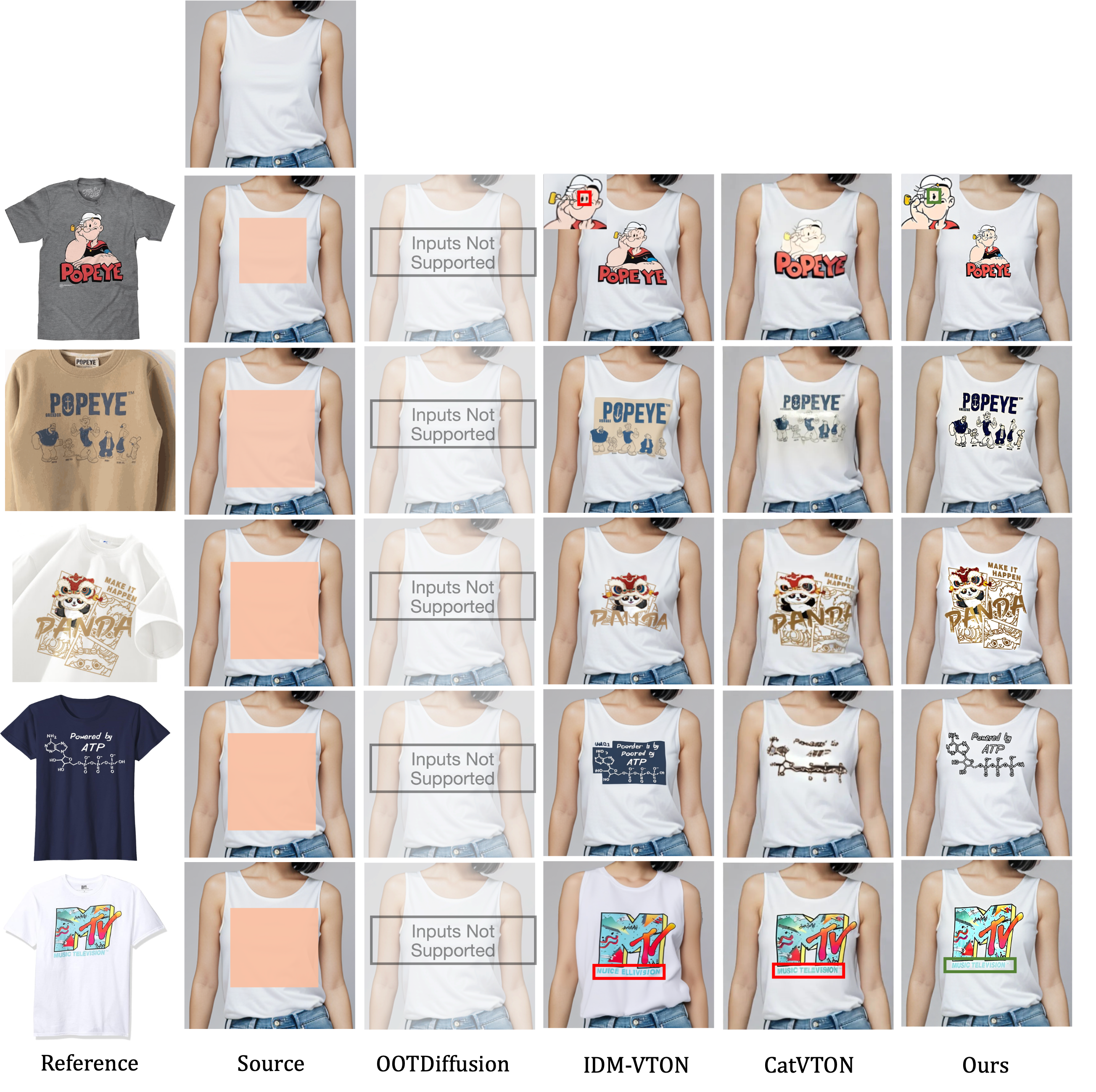}
  \caption{\textbf{Additional Virtual try on (VTO) customization examples.} \textcolor{red}{Red} boxes highlight errors of alternative state-of-the-art models. \textcolor{Green}{Green} boxes highlight correct output from our model. Our method can generate better garment details and layouts. Please zoom in to see details preservation.}
  \label{fig:more_editing}
\end{figure*}

\begin{figure*}[h]
  \centering
  \includegraphics[width=0.85\textwidth]{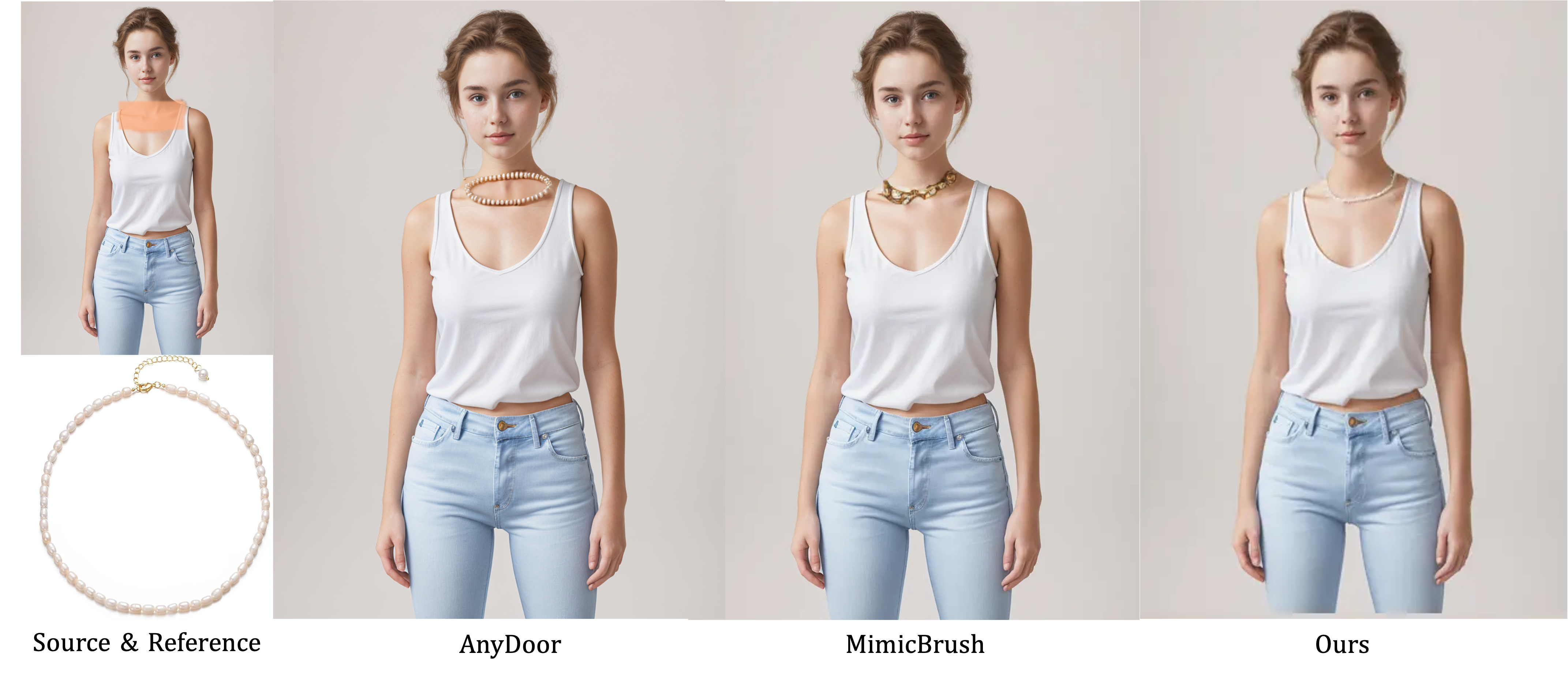}
  \caption{\textbf{Virtual try all (VTA) - Wearable: necklace.}}
  \label{fig:qual7}
\end{figure*}

\begin{figure*}[h]
  \centering
  \includegraphics[width=0.85\textwidth]{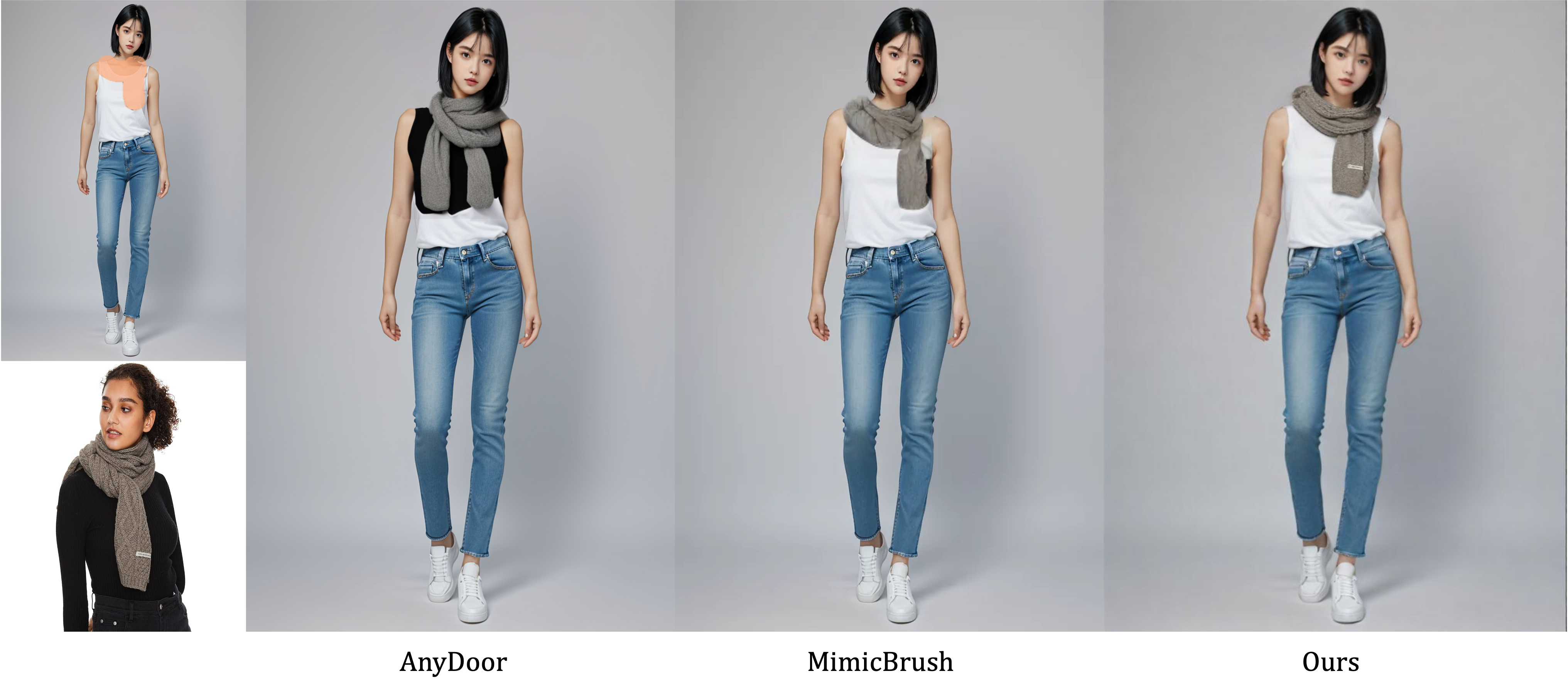}
  \caption{\textbf{Virtual try all (VTA) - Wearable: scarf.}}
  \label{fig:qual8}
\end{figure*}

\begin{figure*}[h]
  \centering
  \includegraphics[width=0.85\textwidth]{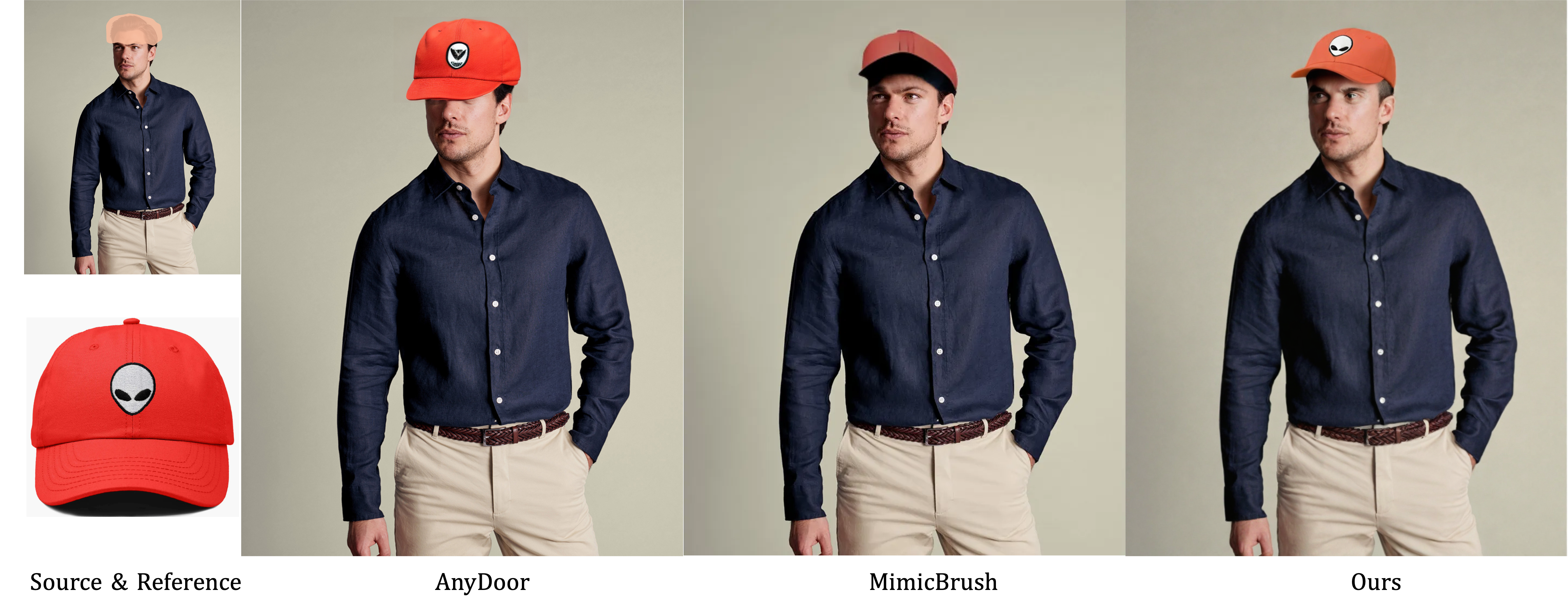}
  \caption{\textbf{Virtual try all (VTA) - Wearable: hat.}}
  \label{fig:qual9}
\end{figure*}

\begin{figure*}[h]
  \centering
  \includegraphics[width=0.9\textwidth]{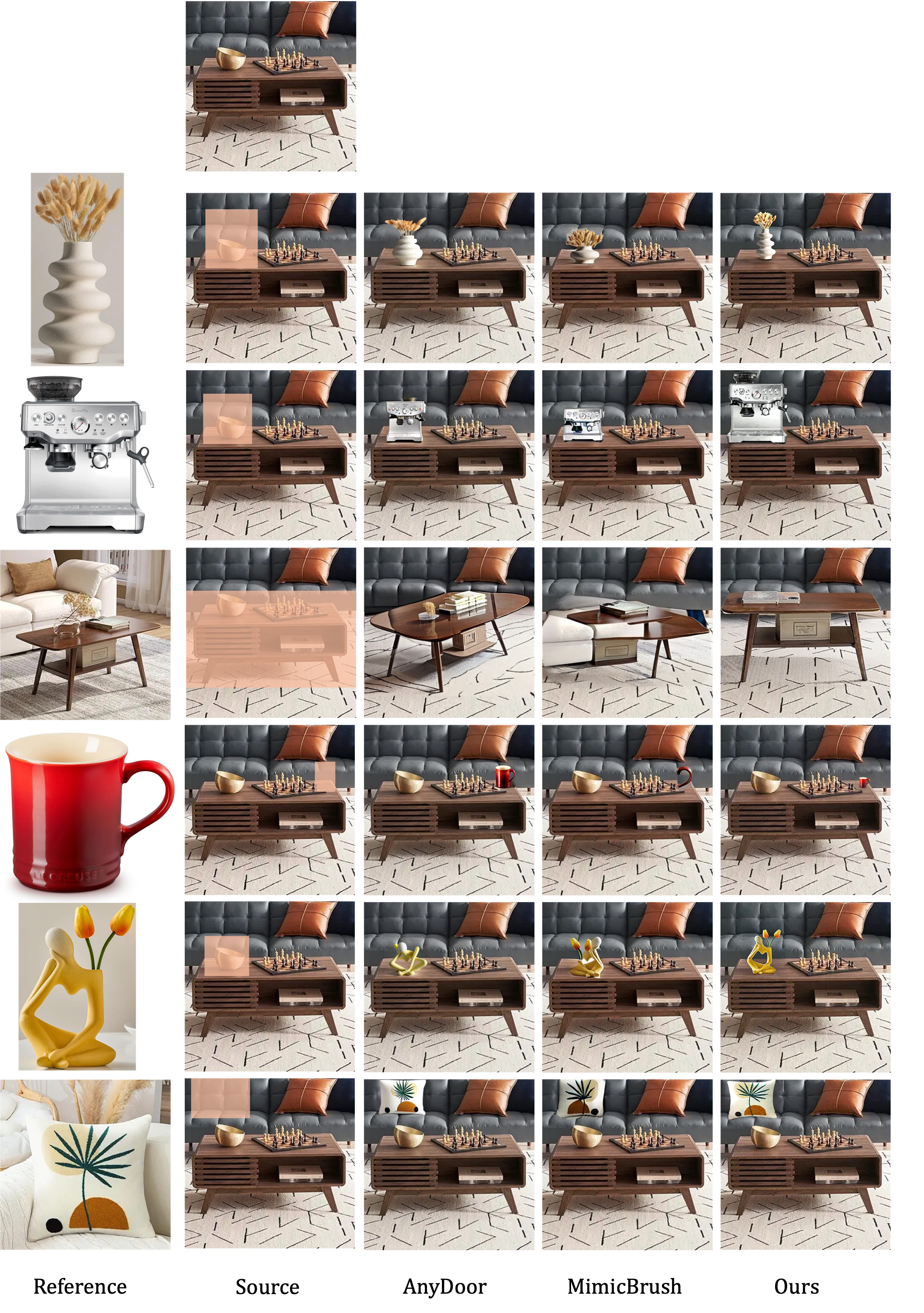}
  \caption{\textbf{Virtual try all (VTA) - Non-Wearable.} Our method can generate better product details and layouts. Please zoom in to see details preservation.}
  \label{fig:more_vta_examples}
\end{figure*}

\begin{figure*}[h]
 \centering
 \includegraphics[width=0.7\textwidth]{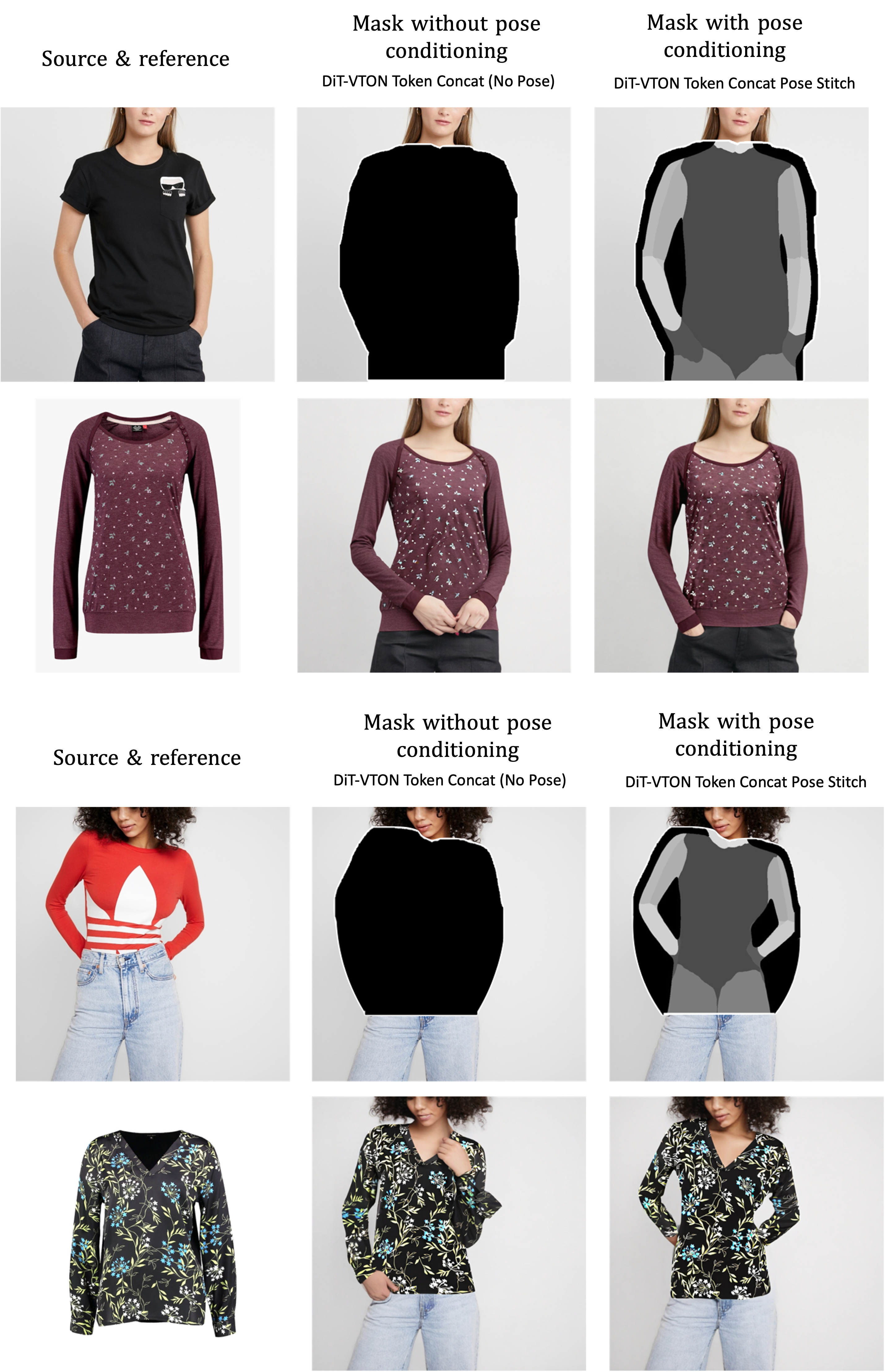}
 \caption{\textbf{In-the-wild try on: pose preservation.}}
 \label{fig:more_pose}
\end{figure*}

We provide qualitative examples of in-the-wild try-on, multiple garments try-on, try-on texture transfer and try-on local editing to compare our models with baselines including OOTDiffusion, IDM-VTON and CatVTON in~\cref{fig:qual1,fig:qual2,fig:qual3,fig:qual4,fig:qual5,fig:qual6,fig:more_vto,fig:more_editing}. We provide qualitative examples of virtual try all to compare our models with AnyDoor and MimicBrush in~\cref{fig:qual7,fig:qual8,fig:qual9,fig:more_vta_examples}. In~\cref{fig:more_pose}, we provide additional virtual try on examples with pose control.

\end{document}